\documentclass{elsarticle}

\usepackage{hyperref}
\usepackage[table]{xcolor}
\definecolor{darkblue}{rgb}{0, 0, 0.5}
\hypersetup{colorlinks=true,citecolor=darkblue, linkcolor=darkblue, urlcolor=darkblue}
\usepackage{amsmath}
\usepackage{tabularx}
\usepackage[autostyle=false, style=english]{csquotes}
\MakeOuterQuote{"}

\newcommand{\secref}[1]{\ref{sec:#1}}

\newcommand{\figref}[1]{\autoref{fig:#1}}
\newcommand{\tabref}[1]{\autoref{tab:#1}}

\newcommand{\tuniquen}[0]{82.3}
\newcommand{\tmemorisedn}[0]{80.0}
\newcommand{\tmemorisedbestofn}[0]{39.8}
\newcommand{\tpiin}[0]{15.4}
\newcommand{\tlogicn}[0]{31.3}
\newcommand{\tfactualn}[0]{46.4}
\newcommand{\tdiscoursen}[0]{52.0}

\newcommand{\tunique}[0]{$\tuniquen{}\%$}
\newcommand{\tmemorised}[0]{$\tmemorisedn{}\%$}
\newcommand{\tmemorisedbestof}[0]{$\tmemorisedbestofn{}\%$}
\newcommand{\tpii}[0]{$\tpiin{}\%$}
\newcommand{\tlogic}[0]{$\tlogicn{}\%$}
\newcommand{\tfactual}[0]{$\tfactualn{}\%$}
\newcommand{\tdiscourse}[0]{$\tdiscoursen{}\%$}

\newcommand{\modeltotal}[0]{nine}
\newcommand{\prompt}[1]{\small{\texttt{#1}}}

\biboptions{sort&compress}

\begin{document}

\title{An Evaluation on Large Language Model Outputs: Discourse and Memorization}

\author[1,2]{Adrian de Wynter\corref{fcor1}}
\ead{adewynter@microsoft.com}
\author[1]{Xun Wang}
\ead{xunwang@microsoft.com}
\author[1]{Alex Sokolov}
\ead{alexsokolov@microsoft.com}
\author[1]{Qilong Gu}
\ead{qilonggu@microsoft.com}
\author[1]{Si-Qing Chen}
\ead{sqchen@microsoft.com}

\cortext[cor1]{Corresponding author.}
\affiliation[1]{organization={Microsoft Corporation},
        addressline={One Microsoft Way},
        postcode={98052},
        city={Redmond, WA},
        country={USA}}
\affiliation[2]{organization={University of York},
    addressline={Heslington},
    postcode={YO10 5DD},
    city={York},
    country={UK}}

\begin{abstract}
We present an empirical evaluation of various outputs generated by \modeltotal{} of the most widely-available large language models (LLMs). 
Our analysis is done with off-the-shelf, readily-available tools. 
We find a correlation between percentage of memorized text, percentage of unique text, and overall output quality, when measured with respect to output pathologies such as counterfactual and logically-flawed statements, and general failures like not staying on topic. 
Overall, \tmemorised{} of the outputs evaluated contained memorized data, but outputs containing the most memorized content were also more likely to be considered of high quality. 
We discuss and evaluate mitigation strategies, showing that, in the models evaluated, the rate of memorized text being output is reduced. We conclude with a discussion on potential implications around what it means to learn, to memorize, and to evaluate quality text. 

\end{abstract}

\maketitle

\section{Introduction}

Large language models (LLMs) are becoming widespread in both research and commercial applications. 
In what is perhaps to date the best example of machine learning democratization, these billion-parameter, resource-intensive models are remarkably easy to use. 
They do not require access to specialized hardware or software, or in-depth understanding of machine learning or natural language processing. 
Instead, their usage is provided through APIs, and are--or claimed to be--able to solve relatively complex tasks with no (zero-shot) or limited (few-shot) prior information. 
This in turn provides a very intuitive and natural way to interact with computers. 

They have also been at the center of a back-and-forth between hype and caution, in no small part due to their poorly-understood capabilities: LLM output tends to be syntactically correct, but sometimes it contains fictitious information \cite{floridi2020gpt}, toxic content \cite{gehman-etal-2020-realtoxicityprompts}, and, of particular interest to this paper, memorized content from the training data \cite{inan2021training,LiangHolistic,carlini2022quantifying}. 

Unfortunately, the ease of access to LLMs is mostly limited to their \emph{use}--that is, submitting text and obtaining responses--and not many models have been made openly available for analysis and vetting by the broader scientific community. 
Models such as GPT-3 \cite{GPT3} and GPT-4 \cite{GPT4} are treated like services, receiving regular updates that may override the work that originally described them.\footnote{In fact, the authors of the latter model have, at the moment, not disclosed any details around the model, except its performance in some tasks.} 
This is both a cause of concern and an opportunity: it is clearly no longer feasible to solely rely on weight analysis and fixed-task corpora to evaluate LLMs. The former since these models may not be openly accessible to the community; and the latter because, since LLM training data is sourced from the internet, it could contain said corpora.

\subsection{Experiment Description}

In this paper we seek to understand to what extent memorization of the training data affects the quality of the output. 
We critically evaluate \modeltotal{} of the better-known LLMs in a way that is meant to be easily reproducible, without needing to access the model's weights or specialized datasets or software. To do this we only use off-the-shelf, readily-available plagiarism detection tools and human annotation. 

The LLMs evaluated in this paper are BLOOM \cite{BLOOM}, ChatGPT \cite{openai_2022}, Galactica \cite{Galactica}, two versions of GPT-3, GPT-4, OPT \cite{OPT}, OPT-IML \cite{OPTIML}, and LLaMA \cite{LLaMA}. 
These models were chosen based on presence in the media, ease of access, and rate of citations on scientific papers one month after their publication and/or announcement. 
These models form a diverse pool: for example, BLOOM is community-built and open-source, while the GPT models are neither. ChatGPT is designed for dialogue, while Galactica is meant for writing assistance in scientific publications. 

For every model, we generated $75$ texts using prompts from $5$ domains: scientific papers, blog posts, knowledge retrieval (KR), long-text autocompletion, and common text openers. 
We then evaluated these texts in four \emph{discourse} categories: presence of personally-identifiable information (PII), factual errors, logical fallacies, and discourse coherence (e.g., grammar and syntax) and two \emph{textual} categories (presence of memorized text and percentage of original text). 

Note that some or all of these categories are called hallucinations in various works. 
In this paper, we consider them separate, and use hallucination only to refer to repeated text. 
Finally, discourse quality in this work is constrained to the aforementioned categories: prose and vocabulary are subjective and we do not evaluate these. 

\subsection{Findings}

Under our evaluation criteria, we find that:

\begin{enumerate}
    \item About \tmemorised{} of the outputs presented some degree of memorization (\tmemorisedbestof{} with a majority vote), with an average of \tunique{} original text. 
    \item About \tfactual{} of the texts generated presented factual errors; \tdiscourse{} had discourse flaws, such as self-contradictory statements; \tlogic{} contained logical fallacies; and \tpii{} generated PII such as incorrect claims and papers misattributed to living scientists. 
    \item These results vary noticeably across models and categories. For example, the GPT-3 models and ChatGPT have the highest incidence of memorization, but the highest-quality discourse. On the other hand, Galactica has one of the highest incidences of outputting factual errors and logical inconsistencies, but this is considerably more prevalent when prompting about blog posts, and not scientific papers.
    \item We find that explicitly prompting the models to avoid outputting memorized content mitigates considerably the amount of content flagged by online plagiarism detection tools.
    \item We find a very strong anticorrelation between factual errors and memorized text for several models, suggesting that the ability of these models to output factual text is related to their ability to memorize it. 
    \item We confirm the relationship between text uniqueness (originality) and memorization, results well-known to the literature \cite{LiangHolistic,carlini2022quantifying}; and extend these results to overall discourse quality, finding that higher-quality discourse is usually returned by the models that memorize the most.
\end{enumerate}

Our work adds to the growing body of literature regarding memorization in LLMs; but also poses two very important questions: \emph{to which extent it is memorization needed for factually-correct, quality output?}, and \emph{what is the amount of memorization needed to perform statistical learning?}. 
Both questions are not new, but are worth revisiting in light of the recent development and adoption of LLMs. 

Finally, we note that LLMs produce very high-quality text, and have many potential uses. 
The concerns about LLMs as tools--for example, in its academic use--are justified, yet invite a deeper reflection around what does it mean to have \emph{quality} text. 
Concretely, we believe that text should now be critically evaluated based in its argumentative and reasoning capabilities, and not on its grammaticality alone. 

We provide multiple supporting arguments and evidence for these questions across our work, and discuss it in depth in Section \secref{conclusion}.

\section{Related Work}\label{sec:relatedwork}
Memorization in machine learning models is known in the literature \cite{carlini2019secret,carlini2022quantifying,inan2021training,LiangHolistic,Aleena}, and has been used to do KR \cite{petroni-etal-2019-language,guu2020retrieval,COMET,carlini2021extracting,lewis2020retrieval,Ziegler}. 
It is also known that certain prompts affects factuality \cite{petroni2020how} and potential to output toxic content \cite{gehman-etal-2020-realtoxicityprompts}. 
\citet{carlini2022quantifying} were the first, to our knowledge, to quantify the amount of memorization done by a language model. 
They found that the model's tendency to memorize text is correlated to the parameter size. 
\citet{PlagiariseLee} found that language models tend to plagiarize in ways distinct to memorization, although their evaluation was limited to smaller language models. 
We note that KR as memorization could be considered a privacy breach, hence we consider the presence PII as one of our discourse flaws. 

We direct the reader to the survey by \citet{liu2023pre} on prompting methods; the work by \citet{ZhouEtAl} on using LLMs to generate prompts; and existing benchmarks such as Piqa \cite{bisk2020piqa}, LAMA \cite{petroni-etal-2019-language,petroni2020how}, and BIG-Bench-Hard \cite{suzgun2022challenging} for evaluating the performance of LLMs across multiple tasks. 
Our work is not designed to be an automated benchmark, and instead focuses on manual analysis of the text. 
It is worth noting that earlier work has shown similar numbers to the ones we report for KR, but for other models and evaluation sets, such as BERT with LAMA \cite{jiang2020can}. 

\citet{LiangHolistic} performed an authoritative evaluation on LLMs prior to the release of some of the models evaluated, to which we point the interested reader. 
Our work is narrower in scope, and focuses on the quality of the output as seen by an user without training, and by the easier-to-access models. 
Our KR findings are largely in line with theirs, although we find the incidence of output of memorized content to be higher. 
We also expand on the correlation between memorization and accuracy, by noting that this extends to overall discourse quality.

Finally, work has been done in characterizing LLMs in terms of fairness and their potential social impact \cite{delobelle2022measuring,domnich2021responsible,chen2022critical,rauh2022characteristics,gehman-etal-2020-realtoxicityprompts,sheng-etal-2019-woman,kurita-etal-2019-measuring}; 
this area is even more critical as they become more accessible. 
A cause of concern is the observation that in some instances LLMs tend to spew nonsense at best, and disinformation at worst, both radicalizing and homogenizing the user base \cite{mcguffie2020radicalization,chen2022critical,gehman-etal-2020-realtoxicityprompts,PlagiariseLee}. 
While we consider potentially harmful content a discourse error, we do not directly study it. 
Whenever applicable, we discuss mitigations put in place by the model creators or API owners.

\section{Methodology}\label{sec:methodology}

\subsection{LLMs Evaluated}\label{sec:llms}

Most of the models evaluated here are transformer-based autoregressive LLMs, trained on large corpora and with parameter sizes in the order of billions. 
The possible exceptions to this are GPT-4 and ChatGPT, whose architectures and other training details are not publicly available. 
With the exception of BLOOM and GPT-4, all are monolingual English models. 
Save for the GPT models, all have a context length of $2,048$ tokens. 
Whenever possible, we use the model's online APIs. Otherwise, we pick the largest model that can fit on a single machine with 8 nVidia V100 GPUs. 
Refer to \secref{llms} for a detailed description of each LLM.

\subsection{Services Utilized}\label{sec:services}

Given that we do not have direct access to the exact data from the models, but all of them rely on crawled websites, we utilize five online services to detect plagiarism given a text. 
All of them, except one, provide the percentage of unique text, and we average this metric for reporting purposes. 
We do not inquire as to the inner workings of each service and, to avoid overloading the services, only one of the authors collects the output by manually submitting each of the model outputs. 
Since our goal is not to verify the quality of each service, the results are reported in aggregate and we do not perform any comparisons of their relative performance. %

\subsection{Prompting and Querying Guidelines}\label{sec:prompting}

Whenever possible, we specify the token output length to $2,048$ tokens. 
For submission, we removed hallucinations. 
We do not remove non-English text. 
For the services that require a word limit, we truncate the text to match. 
When the submission does not meet the word limit (e.g., a word or a simple sentence), we consider it $100\%$ unique text, and $0\%$ memorized text. 
Unless indicated otherwise, our generation relies on the API's default parameters. 

We follow a best-of-five approach, calling the model and picking the longest (post-hallucination cleaning) text generated. 
In the case of ChatGPT, we also attempt to query the model until we obtain an answer instead of the canned responses requesting for more information. 
We do not request knowledge prior to 2020, since most models have a cutoff sometime around 2021. 
Detailed guidelines can be found in \secref{methodologyexp}. 

\section{Text Analysis}\label{sec:syntacticanalysis}

\subsection{Domain Prompts}

We hand-crafted or crawled  15 prompts for each of our five domains. The crawled prompts are common texts that are more likely to appear in the training text, and belong to the \textsc{Long text} and \textsc{baseline} domains. The hand-crafted prompts are either based on common topics for \textsc{blog posts} or \textsc{scientific articles}, or hand-crafted yet easily-verifiable through Wikipedia (a dataset common to all models), questions for the \textsc{KR} domain. 

It is worth mentioning that \textsc{long text} is considered a text completion task with long context, as opposed to \textsc{baseline}. 
This distinction becomes clear when determining memorized content: \textsc{baseline} should contain some plagiarized content (we do not remove the prompt from the text when evaluating) and is meant to calibrate the effectiveness of the services. \textsc{long text} is simply a text completion task: the prompt is removed ahead of measuring memorization. 
We consider both tasks \emph{conditioned} prompts, however: the prompt contains existing text and the model should not autocomplete the rest of the text.

Although not a strict baseline, \textsc{baseline} allows us to quantify an LLM's propensity to output memorized content from conditioned prompts, in addition to the sensitivity of our services to a verifiable, common ground truth. 

See \tabref{prompts} for sample prompts per domain and the appendix for further samples. 

\begin{center}
\begin{table}[h]
\begin{tabular}{|c | m{0.68\columnwidth} |} \hline
Domain & Sample Prompt  \\ \hline
\textsc{Blog posts} & \prompt{Growing up in Leeuwarden} \\ \hline
\textsc{Scientific papers} & \prompt{Large language models are}\\ \hline
\textsc{KR} & See \secref{samplekr} \\ \hline
\textsc{Long text} & See \secref{samplelongtext} \\ \hline
\textsc{Baseline} & \prompt{It was the best of times, it was the worst of times, it was} \\ \hline
\end{tabular}
\caption{Sample prompts for the domains. 
\textsc{Baseline} (bottom) is a well-known string that occurs frequently at the beginning of texts, and corresponds to \emph{A Tale of Two Cities} by Charles Dickens. 
For \textsc{long text}, we chose famous speeches and articles, but unlike \textsc{baseline}, we start and truncate them somewhere in the middle of the text.}\label{tab:prompts}
\end{table}
\end{center}

\subsection{Evaluation Criteria}

We calibrated our services with the \textsc{baseline} domain, containing common openers for texts, both literary (\tabref{prompts}) and informal, from web-crawled corpora ("copy pasta"). 
In addition to evaluating memorization and text uniqueness through the services, we perform human annotation of the text under the following categories:
\begin{enumerate}
    \item Presence of PII. This includes websites, emails, and names of living people to whom work or claims are misattributed. 
    \item Factual errors: claims in the text that cannot be verified by peer-reviewed and trusted sources. This includes incorrect answers to the \textsc{KR} domain. 
    \item Logical inconsistencies: presence of fallacies in the text, such as incorrect proofs and self-contradicting statements. We make an exception for rhetorical devices when their use is clearly intentional (e.g., antitheses such as "\prompt{It was the age of wisdom, it was the age of foolishness}"). 
    \item Discourse inconsistencies: grammatical errors, ill-formatted text, and poor argumentation in an informal-logic context. Note that fluent discourse that may be nonsensical is not considered a discourse inconsistency.
\end{enumerate}

See \secref{methodologyexp} for more fine-grained rules and explanations of each of the categories; and \secref{samplesyntax} for a sample of the evaluation process.

\subsection{Main Results}\label{sec:mainresults}

We present the main results in \tabref{mainresults}. 
Overall, we found that GPT-3.51 presented the highest-quality discourse (only $18.7\%$ flaws), but also had one of the highest incidences of memorized content ($89.3\%$). 
The highest proportion of logical errors ($55\%$) was done by OPT-IML, while Galactica presented the largest amount of factual errors ($73.3\%$). 
 See \secref{fullbreakdown} for a full breakdown of the results over every domain and model. 

We compared these results with and without conditioned prompts (the ones that contain memorized text, \textsc{baseline} and \textsc{long text}). 
Removing \textsc{baseline} from our analysis did not have a noticeable impact on neither discourse quality ($+0.6\%$) or percentage of memorized text ($-3.5\%$). 
When also removing \textsc{long text}, the unique text percentage drops noticeably ($-19.4\%$), although the discourse quality improves ($-12.7\%$ fewer discourse flaws). 

In terms of dialectical quality, ChatGPT had the second-lowest incidence of factual errors, as well as the lowest incidence of logical errors ($7.6\%$). 
Remarkably, it was also able to achieve a $0.0\%$ PII error rate. 
While other models, such as BLOOM, output anonymized tokens (e.g., "PII\_EMAIL", instead of an email address), they do not appear to achieve this consistently. 
We also observed some of the other callouts from \secref{llms}, such as OPT and GPT-4's tendency to output dialogues (such as Reddit threads). 

\begin{center}
\begin{table}[h]
\centering
\begin{tabular}{| l || m{0.145\columnwidth} | m{0.1\columnwidth} || m{0.09\columnwidth} | m{0.085\columnwidth} | m{0.085\columnwidth} | m{0.11\columnwidth} |} \hline
Model & Memorized ($\downarrow$) & Original ($\uparrow$) & PII ($\downarrow$) & Logical ($\downarrow$) & Factual ($\downarrow$) & Discourse ($\downarrow$) \\ \hline
BLOOM & \cellcolor{blue!10}\textbf{53.3/22.7} & 91.5 & 14.7 & 45.3 & 49.3 & 73.3 \\ \hline 
ChatGPT  & 87.8 / 37.3  & 71.7 & \cellcolor{blue!10}\textbf{0.0} & \cellcolor{blue!10}\textbf{7.6} & 37.9 & 24.2 \\ \hline 
Galactica& 72.0 / 26.7  & 92.0 & 31.1 & 52.0 & 73.3 & 77.3 \\ \hline 
GPT-3.5  & 86.7 / 56.0  & 73.0 & 16.0 & 20.0 & \cellcolor{blue!10}\textbf{34.7} & 45.3 \\ \hline 
GPT-3.51 & 89.3 / 56.0  & 71.2 & 5.3 & 13.3 & 40.0 & \cellcolor{blue!10}\textbf{18.7} \\ \hline 
GPT-4    & 94.7 / 58.7  & 73.8 & 9.3 & 17.3 & \cellcolor{blue!10}\textbf{34.7} & 36.0 \\ \hline 
LLaMA    & 82.7 / 58.7  & 78.1 & 9.3 & 17.3 & 36.0 & 56.0 \\ \hline 
OPT      & 65.3 / 26.7  & \cellcolor{blue!10}\textbf{93.0} & 25.3 & 40.0 & 48.0 & 78.7 \\ \hline 
OPT-IML  & 74.7 / 34.7  & 91.9 & 21.3 & 54.7 & 53.3 & 62.7 \\ \hline\hline
Total    & 80.0 / 39.8  & 82.3 & 15.4 & 31.3 & 46.4 & 52.0 \\ \hline 
Total (*)& 76.5 / 35.9  & 82.6 & 15.6 & 28.1 & 51.0 & 52.6 \\ \hline 
Total ($\dagger$) & 58.7 / 25.7  & 62.9 & 11.1 & 20.8 & 42.6 & 39.3 \\ \hline 
\end{tabular}
\caption{Main results of our analysis. Best-performing model in each metric is bolded. Results without \textsc{baseline} are marked with an asterisk (*); results without \textsc{baseline} and \textsc{long text} are marked with a dagger ($\dagger$). 
For memorized text we report both the incidence when at least one service flagged the text (left) and as a majority vote (right). 
Overall, GPT-3.51 presented the highest-quality discourse, but it also presented the second-highest incidence of memorized content. The highest number of factual, PII and logical errors were performed by Galactica, OPT and OPT-IML, respectively.}\label{tab:mainresults}
\end{table}
\end{center}

\subsection{Variable Correlation} 
\begin{figure}[h]
\centering
\begin{minipage}[t]{.49\textwidth}
  \centering
  \includegraphics[width=0.9\linewidth]{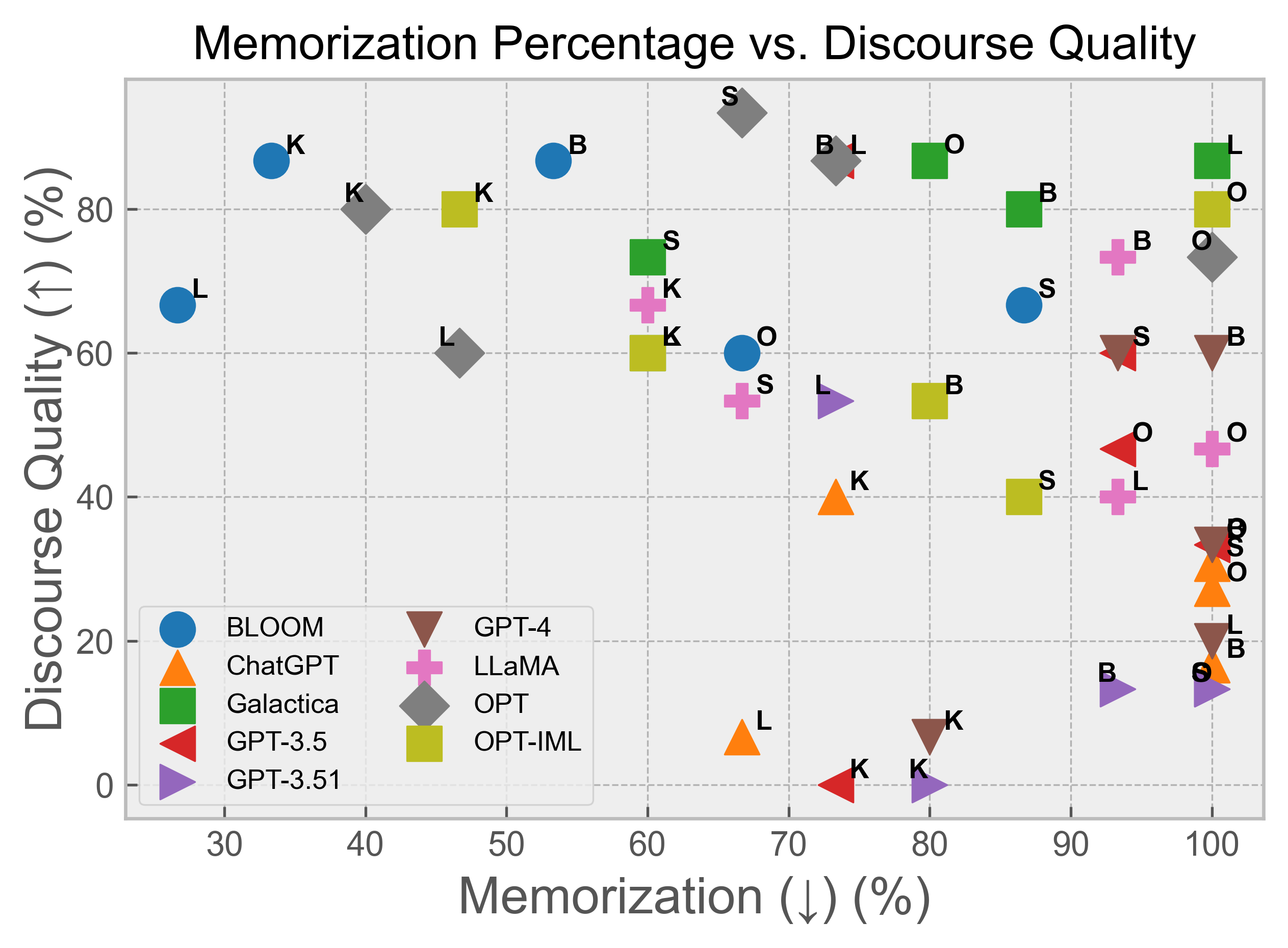}
  \caption{Discourse quality as a function of the percentage of memorized text. 
  The text output quality was higher when the model presented higher proportions of memorized content (e.g., GPT-3.51). 
  This plot does not account for dialectical errors. The domains evaluated \textsc{baseline}, \textsc{blog post}, \textsc{long text}, \textsc{scientific papers}, and \textsc{KR}, are denoted by $O$, $B$, $L$, $S$, and $K$, respectively.}
  \label{fig:discoursemem}
\end{minipage}\hfill
\begin{minipage}[t]{.48\textwidth}
  \centering
  \includegraphics[width=\linewidth]{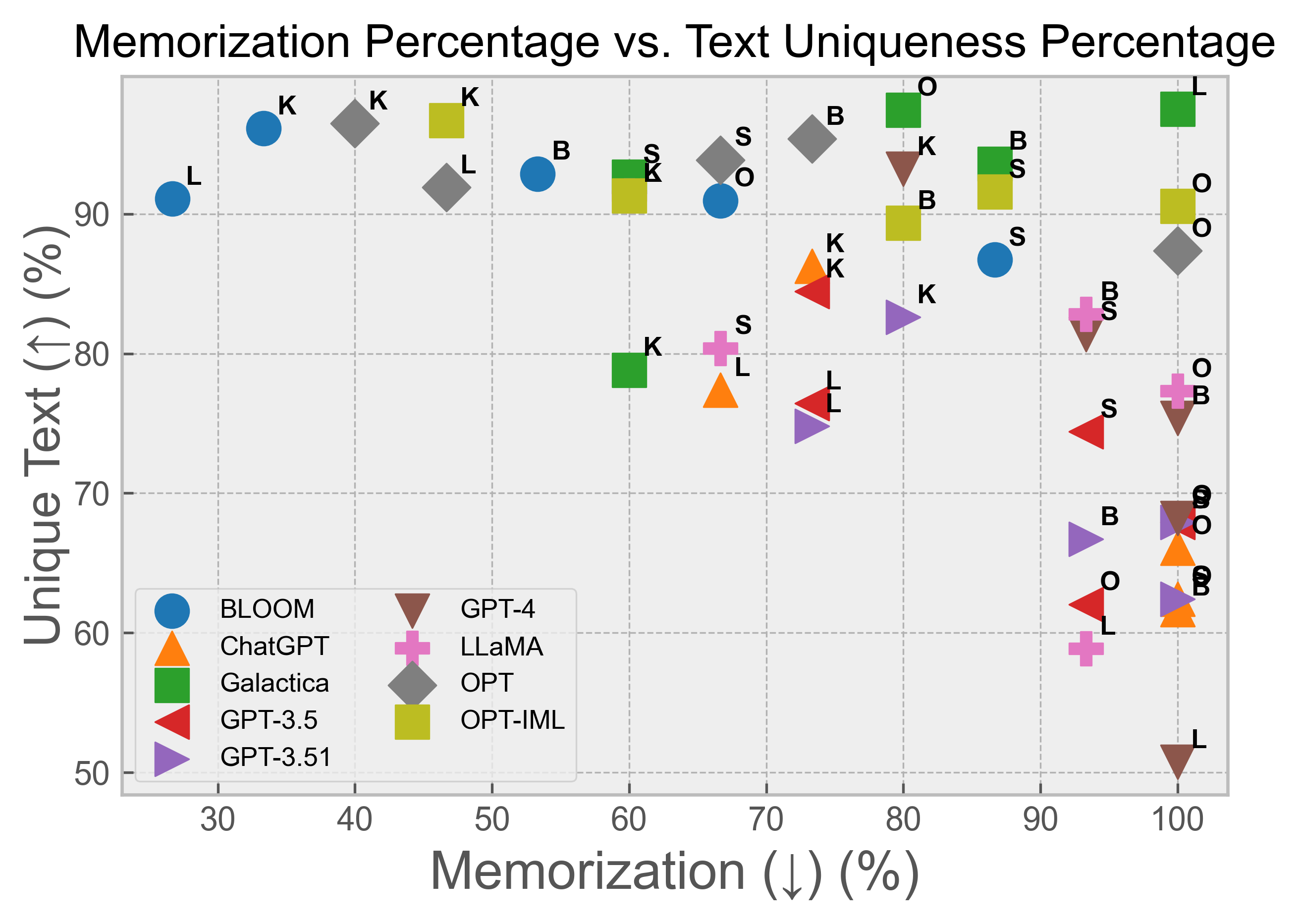}
  \caption{Unique (original) text percentage as a function of the percentage of memorized text. 
  The uniqueness of the text returned dropped as a function of the amount of memorized content flagged by the services. This plot does not account for discourse quality or dialectical errors. The domains evaluated \textsc{baseline}, \textsc{blog post}, \textsc{long text}, \textsc{scientific papers}, and \textsc{KR}, are denoted by $O$, $B$, $L$, $S$, and $K$, respectively.}
  \label{fig:memuniqueness}
\end{minipage}
\end{figure}

We compared the correlation between the memorized text and other variables, such as discourse quality (\figref{discoursemem}) and original text (\figref{memuniqueness}), and evaluated them with Pearson's $\rho$ correlation coefficient. 

On average, there is a weak anti-correlation ($\rho=-0.40$) between the quality of the output text and the percentage of memorization done. However, the amount of unique text being output is more strongly anti-correlated with the memorized text ($\rho=-0.61$) and the discourse quality ($\rho=-0.63$). 

We found a weaker correlation, on average, between the factuality of the output and the memorized text ($\rho=-0.45$). 
When looking at the breakdown per model, however, we notice a much wider variability across models. 
For example, the quality of the output was very strongly correlated with the text uniqueness for Galactica ($\rho=0.97$) but not for GPT-3.51 ($\rho=0.0$). See \tabref{rhos} for a breakdown over metrics. 

This suggests that the interplay between discourse quality, factual errors, and text uniqueness varies heavily across models.

\begin{center}
\begin{table}[h]
\centering
\begin{tabular}{| l || c | c | c || c |} \hline
Model     & Discourse & Factual & Original & Original ($\dagger$) \\ \hline
BLOOM     &  -0.40 & -0.29 & -0.74 & 0.72\\ \hline 
ChatGPT   & 0.20 & -0.19 &  \cellcolor{blue!10}\textbf{-0.89} & 0.23 \\ \hline 
Galactica & \cellcolor{blue!10}\textbf{0.81} &  \cellcolor{blue!10}\textbf{0.58} &   \cellcolor{blue!10}\textbf{0.70} & \cellcolor{blue!10}\textbf{0.97} \\ \hline 
GPT-3.5   &  0.00 & -0.33 &  -0.78 & \cellcolor{blue!10}\textbf{-0.28} \\ \hline 
GPT-3.51  &  \cellcolor{blue!10}\textbf{-0.53} & -0.80 & -0.81 & 0.0 \\ \hline 
GPT-4     & 0.51 & \cellcolor{blue!10}\textbf{-0.98} &  -0.80 & 0.02 \\ \hline 
LLaMA     & -0.35 & -0.48 &  -0.59 & 0.80 \\ \hline 
OPT       & 0.17 & -0.79 &  -0.71 & 0.47 \\ \hline 
OPT-IML   & -0.22 & -0.78 &  -0.74 & 0.47 \\ \hline\hline 
Average   & -0.39 & -0.45 &  -0.61 & 0.63 \\ \hline 
\end{tabular}
\caption{Correlation of memorized text percentage with respect to discourse quality, factual errors, and text uniqueness, as measured with Pearson's correlation coefficient. We also include correlation of discourse quality with respect to text uniqueness, marked with a dagger ($\dagger$). 
We highlight the largest correlated and anti-correlated models. 
GPT-4 has a very strong anti-correlation between memorized text percentage and factual errors. 
GPT-3.5 reports no correlation between discourse quality and percentage of memorized text, in spite of it having flagged 89\% of its outputs as memorized. However, it can be seen that text uniqueness is strongly anti-correlated with percentage of memorized text. Overall Galactica showed positive correlations between text uniqueness, factual errors, and memorized percentage.}\label{tab:rhos}
\end{table}
\end{center}

\subsection{Prompt Length and Memorization}

\begin{figure}[h]
\centering
\includegraphics[width=0.95\textwidth]{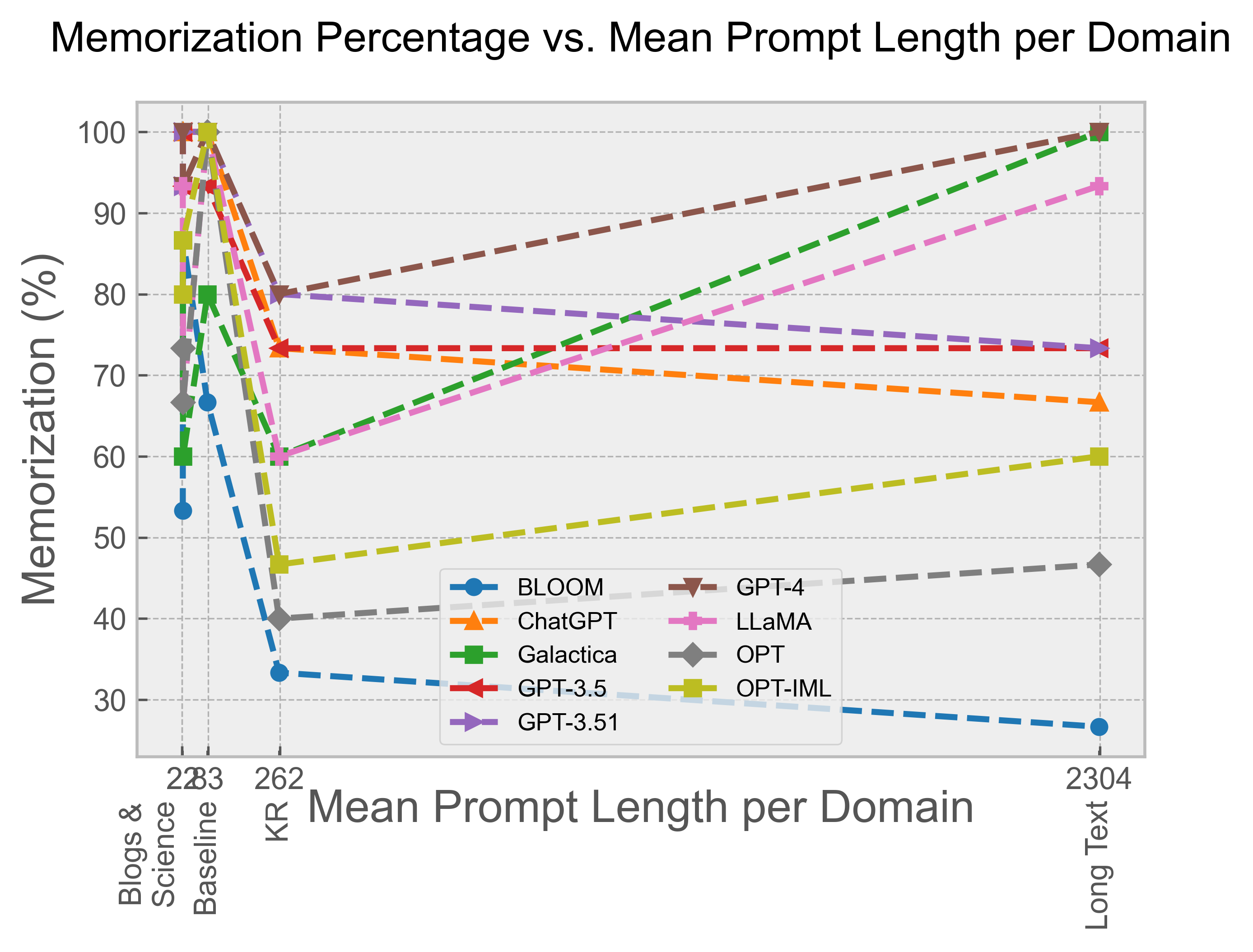}
    \caption{Memorization percentage as a function of the length of the prompt (in characters), per domain. 
    Shorter prompts caused the models to output more memorized content. The general trend of the models to output memorized content with respect to the prompt length repeated itself over all the models evaluated, suggesting that this is a behavior intrinsic to LLMs.}
    \label{fig:promptlenmem}
\end{figure}

We compared prompt length and amount of memorization in \figref{promptlenmem}. 
We found that shorter prompts, not accounting for \textsc{baseline}, are related to higher proportions of memorization, and this pattern is consistently reflected over the models. 
This also suggests that the tendency to output memorized text is general to all LLMs evaluated.

\subsection{Mitigation Strategies for Memorization}
\begin{sloppypar}
We attempted to mitigate outputting memorization by directly prompting the models to not output pre-existing content. \citet{PengEtAl} described a similar strategy to mitigate counterfactual content generation. 
Our prompt is: \prompt{You are a creative writer, and you like to write everything differently from others. 
Your task is to follow the instructions below and continue writing at the end of the text given}. The instructions (given in markdown format) are "\prompt{Write in a way different from the actual continuation, if there is one}", and "\prompt{No plagiarism is allowed}". 
\end{sloppypar}

We tested this approach with our \textsc{long text} domain, and with ChatGPT, GPT-3.51 and GPT-4. The domain was chosen because it contained conditioned prompts; the models were selected based on their discourse-memorization trade-off. 
The results are in \tabref{mitigation}. 
Overall we observed that all models were receptive to this prompting strategy, and showing a marked decrease on the amount of memorized content being flagged by the services. ChatGPT showed an average increase of $+29\%$ original text, while GPT-4 decreased in its memorization incidence by $-40\%$. 
An outlier was GPT-3.51, which showed an increase in memorized content being flagged by at least one service ($+7\%$) but lower in a best-of-three context ($-57\%$) and higher original content output ($+2\%$). 

\begin{center}
\begin{table}[h]
\centering
\begin{tabular}{| l || c | c || c | c |} \hline
Model & Memorized ($\downarrow$) & Original ($\uparrow$) & Memorized$^\dagger$ ($\downarrow$) & Original$^\dagger$ ($\uparrow$) \\ \hline
ChatGPT  & 66.7 / 33.3 & 70.6 & 60.0 / 6.7 & 99.1  \\ \hline
GPT-3.51 & 73.3 / 60.0 & 79.1 & 80.0 / 13.3 & 81.0  \\ \hline
GPT-4    & 100.0 / 86.7 & 53.4 & 60.0 / 26.7 & 68.8   \\ \hline
Total    & 73.3 / 47.5 & 84.8  & 66.7 / 15.6 & 83.0  \\ \hline
\end{tabular}
\caption{Memorization mitigation results. 
For this experiment we report the results from \textsc{long text} (\secref{longapp}). 
Our results from the mitigation analysis are marked with a dagger ($\dagger$). 
Overall all models were receptive to this strategy: ChatGPT presented the largest improvement in original text percentage ($+29\%$), and GPT-4 had the largest decrease of content flagged as memorized ($-40\%$) overall.}\label{tab:mitigation}
\end{table}
\end{center}

\subsection{Discussion}\label{sec:discussiondoc}

In \secref{mainresults} we mentioned that removing both \textsc{baseline} and \textsc{long text} had a noticeable effect on both mean original text and discourse quality. 
We found these variables to be correlated to percentage of memorization. 
Solely removing \textsc{baseline} did not alter noticeably the numbers we obtained. 
These observations suggest that a considerable amount of the discourse quality was contributed by both biased prompts--and hence by the memorized text. 
This means that, although it is relatively easy to cause LLMs to output the training set, it is much more likely to be output if the prompt is conditioned (i.e., biased) to do so (\tabref{mainresults}). 
This is perhaps not too surprising as models are meant to output the most likely sequence of tokens being observed; unless the dataset has been properly deduplicated, it will be the remainder of the conditioned text. 

When comparing \textsc{baseline} and \textsc{long text}, we can still observe a--perhaps weak--linear correlation between prompt length and percentage of memorized text. 
This suggests that memorization is a phenomenon not solely intrinsic to the syntactical content of the prompt, but also to the models themselves. 
The rest of \figref{promptlenmem} supports this hypothesis, as the tendency to memorize varies (almost) analogously over all LLMs. 

A recent result by \citet{carlini2022quantifying} mentions that the models that memorize the most are the largest. 
We observed that, although BLOOM, GPT-3, and (perhaps) ChatGPT have about the same number of parameters, their memorization rates are considerably different. 
This suggests that propensity to memorize content is a process tied to the architecture and the training process. 
We discuss this further in \secref{conclusion}.

\section{Limitations}\label{sec:limitations}
Our study is limited by the relatively few number of prompts and models evaluated. 
Certain LLMs, like PaLM \cite{PALM}, are not available at all, and hence did not meet our selection criteria. 
Other models, such as GPT-NeoX \cite{NeoX}, were not considered simply due to a ranking matter. 
Further studies should certainly evaluate these valuable, open-source LLMs. 

The number of prompts is certainly a concern, although it is an intrinsic downside when performing human annotation, especially at the text length and critical reading required by our study. 
We argue that such a relatively small number of prompts is still representative of the broader issues of this technology, given how frequently these showed. 
The services also have their own limitations: they didn't seem to be able to detect text that had been typeset, even when some were clearly taken from online sources. 

It is worth mentioning that our comparison is perhaps not entirely fair. The nature of the operation for the GPT models is opaque by design. 
We cannot be fully sure that we have compared the bare models directly, and any other upstream/downstream components could have played a factor in our results.

\section{Conclusion}\label{sec:conclusion}

In this work we explored the relationship between discourse quality and memorization for LLMs. 
Our conclusions are somewhat unsatisfactory: some of these models are able to output high-quality text, but often are flagged as memorized by our services. 
On the other hand, some models did not show a high incidence of memorized content, at the expense of returning text that could require considerable effort to clean and parse. 

We also pointed that, while factuality, discourse quality, and text uniqueness are often correlated, it varies across models: GPT-4 has a very strong anti-correlation between factual errors and memorized text, but Galactica--the model that had the highest amount of factual errors on average--presents a weak correlation instead. 
Finally, we showed that by explicitly prompting the model to not output memorized content we could obtain a consistently much lower amount of text being flagged as such.

It is quite possible that our findings are simply a training issue, and not a matter of overparametrization: if the data has not been deduplicated; and/or it is encouraged to return the most likely sequence of tokens (out of many non-distinct sequences), the model would simply encode most likely sequences: the ones it has already seen. 
This hypothesis is supported by the work by \citet{Chinchilla} from a parameter perspective, by \citet{pmlr-v162-borgeaud22a} from a data-size perspective, and by \citet{pmlr-v162-kandpal22a} and \citet{lee-etal-2022-deduplicating} from a deduplication perspective, as well as the findings from this paper. 

That being said, a model that has learned (not memorized) a properly-deduplicated, evenly-distributed, sufficiently-representative dataset would (in theory) not have a correlation between text uniqueness, memorized content, and discourse quality \cite{su2022contrastive}. 
In other words, \emph{learning} such a distribution would imply having a lower chance of being flagged by the services; 
and hence be likelier to output higher-quality text. 
Yet, it has been shown that learning does require some amount of memorization \cite{Feldman}. 

However, if a model is able to output verbatim Kennedy's "We choose to go to the Moon" speech from a few tokens, does that constitute learning, or memorization? 
The model has, in fact, encoded correctly the training set--but solely because "\prompt{We choose to go to the moon in this decade, and do the other things...}" is a very commonly occurring string. 
Then again, for all intents and purposes, verbatim output of existing text \emph{is} memorization. 

It is clear from our work that the models evaluated here show a direct link between the quality of the content generated, and the amount of memorized text. 
But, to what extent is it possible to say that these models are learning? 
Is this memorization beneficial, and when does it become a liability \cite{Parrot}? 

Finally, the amount of memorization and discourse flaws presented by these models invite for a deeper reflection on how does the scientific community evaluate model performance. 
Many models generated consistently grammatical text, but with errors in both formal and informal reasoning. 
This ties both to the recent claims about these models achieving artificial general intelligence (AGI); and to their broader argumentative capabilities. 
We can see that these LLMs do not consistently withstand a critical evaluation of their output. 
This is both a point of improvement for research (how can we automate these measurements without fully relying on statistical methods?) and an area to keep in mind so that LLMs are used safely, responsibly, and with the benefit of the user in mind.

\appendix

\section{LLMs Evaluated}\label{sec:llms}
\subsection{BLOOM}

BLOOM is a multilingual model trained to work in $46$ natural and $13$ programming languages. 
Unlike the other models in this paper, it is community-sourced, and developed by volunteers. 
In the API \cite{BLOOMAPI} the authors indicate that the model may contain PII, harmful content, incorrect statements, and skew and influence users. 
Compared to OPT-175B, the authors indicate that the model performs considerably better in human evaluation tasks. 

Architecturally, it is a modified Megatron-LM GPT2, and trained with cross-entropy and mean reduction. The authors made an explicit attempt to leverage the internet to build the training corpus, sourced primarily ($38\%$) from OSCAR \cite{OSCAR1,OSCAR2}, a subset of the Common Crawl.\footnote{\url{https://commoncrawl.org/}} 
In this paper we evaluate version 1.3, or the checkpoint corresponding to July 6 through the API. 
For KR, we follow the author's recommendation and not attempt to chat with it, but simply paste the query. 

\subsection{ChatGPT}

ChatGPT \cite{openai_2022} is a model especially trained for dialogue with reinforcement learning with human feedback \cite{InstructGPT}. 
Little is known in detail about this model, outside of the post in the reference, although the versions we tested were stated to be GPT-3 based. 
The model is a service and constantly updated. 
We test the version from January 9, 2023. 
The authors claim that it has fewer incidences of counterfactual errors when compared with GPT-3. 
They also point out that this model is prompt-sensitive, and may return answers with discourse errors ("nonsensical answers") \cite{openai_2022}.
For prompting this model, we simply paste our prompts. 

\subsection{Galactica}

Galactica is a model designed for scientific article writing assistance, based off OPT. 
Technically, this model supports both English and the LaTeX markup language. 
The training corpus is collected from open-access scientific papers and other reference materials (e.g., encyclopedias), and tokenized to support citations, reasoning, sequences, and other article-specific syntactic structure. 
The model has been prompt-pretrained. 
The authors mention that the model is less toxic and biased than other LLMs; 
and that it is able to answer general-knowledge questions, albeit at a lower performance. 
They include prompting instructions for both scientific-article generation and question answering. 
We follow their instructions, and utilize the \textsc{large} (30B) version of the model.

\subsection{GPT-3}\label{sec:gpt3}
In this paper, we evaluate the \textsc{text-davinci-002}  and \textsc{text-davinci-003} models, through the API \cite{OAIPlayground}. 
There is no information about parameter size for these models, but the original \textsc{davinci} model (GPT-3) is stated to be 175B parameters \cite{OAIModelIndex}. 
The \textsc{text-davinci-002}  and \textsc{text-davinci-003} models are also often referred as GPT-3.5 and GPT-3.51, respectively. 
They are based off InstructGPT--that is, they are instruction-pretrained \cite{InstructGPT,OAIModelIndex}. 
These models are trained on a custom-filtered and deduplicated version of the Common Crawl, with extra data added to improve its quality, and optimizing for the cross-entropy loss. 
The \textsc{text-davinci-003} model is considered an updated version of \textsc{text-davinci-002}, with newer data. 

The authors indicate that this model has semantic repetition at the document level, and presents a tendency to contradict itself and insert unrelated paragraphs. 
They indicate as well that GPT-3 has biases. The API contains a content moderator that flags harmful text. 
For prompting this model, we follow the documentation, but we do not provide a ground truth \cite{OAIPrompt}.

\subsection{GPT-4}\label{sec:gpt4}
There is very little information on GPT-4 outside of its performance in custom benchmarks. 
No details around architecture, training procedure, or model size are specified. 
The authors indicate in the technical report that this model makes reasoning and factual errors, although at a lower rate than the GPT-3 models. 
Unlike other GPT models, GPT-4 is explicitly mentioned to be multilingual and multimodal. 
To ensure a fair evaluation with the other models, we limit the output to $2,048$ tokens. 

\subsection{LLaMA}\label{sec:llama}

LLaMA is a family of pre-trained language models designed to perform well during inference. 
The training set is all publicly available, and is composed primarily of CommonCrawl (67\%) and C4 \cite{RaffelC4} (15\%) data. 
A portion (2\%) of the corpus comes from Stack Exchange, and we observed significant verbatim output from this website. 
The authors indicate that the largest (65B) model is competitive with much larger models, such as GPT-3. 
They attribute this to their optimized pretraining strategy. 
This model is not instruction-finetuned. 

The authors evaluated toxicity and other biases and found them to be slightly better than OPT-175B and GPT-3. They observed, however, that these issues scale with model size. In this paper we evaluate the 65B version of LLaMA and do not specifically follow any prompting strategies. 

\subsection{OPT}\label{sec:opt}

OPT is a family of open-source models akin to GPT-3. 
The corpora utilized are all publicly available, and include parts of the Pile \cite{ThePile}, itself a super set of Wikipedia and Common Crawl; and Pushshift.io \cite{Pushshift}, which includes Reddit threads as conversational sets (i.e., flattened chains of comments). 

The authors indicate that OPT-175B, the largest model in the class, is ineffective when the prompt is a question. 
They note that this causes the model to generate dialogue, counterfactual statements, harmful content, and repetitive output. 
We follow the author's indications that sampling mitigates this problem, and also generate various prompts as described in \secref{prompting}. 
We utilize the 30B version, as it is the largest that would fit in our GPUs without extensive modification.

\subsection{OPT-IML}\label{sec:optiml}
OPT-IML are a family of models analogous to OPT, but instruction pre-trained for better performance. In this paper, we consider the 30B OPT-IML-Max version, which has been trained in approximately $2,000$ NLP tasks. 
The authors indicate that this line of models have the same risks associated with other LLMs.

\section{Text Evaluation Criteria}\label{sec:methodologyexp}

\subsection{Discourse Errors} We separate discourse errors into two major categories: syntactical errors (e.g. grammar) and semantic errors. 
For syntactical errors, we noticed that most models have a very low incidence of misspelled words and malformed sentences. 
However, BLOOM and OPT often output code or web-scrapped content such as HTML tags. 
We consider this a syntactical failure. 
For GPT-4 specifically, the model tends to output repetitive boilerplate ("there are many ways to answer this") as well as answer a question with multiple answers, some of which contradicted one another. 
We did not consider the boilerplate a discourse error, and for \textsc{KR} we only took into account the first answer. 

Semantic errors are any flaws in the content of the discourse. 
This may involve, for example, not answering what it is prompted on, not staying on topic, or outputting content from websites (e.g., discussion forums and dialogue) when not requested. 
We consider all of this semantic failures.

We do \emph{not} consider a discourse error a grammatically-correct text that may not be considered well-structured. 
For example, in several instances OPT generated stories with a beginning and no end, where it just looped over the previous structure but altering it. 
Since this is a relatively common literary device, we do not consider it a flaw.

\subsection{Factual Errors} We consider factuality to be global. 
Any claims that the model make are fact-checked through either peer-reviewed papers or trustworthy sources. 
In the case there is a claim that could have changed over time (for example, claiming that a cloud provider supports $200$ countries), we utilize the Internet Archive's Wayback Machine\footnote{\url{https://web.archive.org/}} for fact-checking instead. 
It is worth noting that in multiple instances this yielded cases of plagiarism otherwise not detected by the services, and hence not reported in our final numbers. 
Unverifiable facts (e.g., the eating habits of hobbits) are not considered a factual error.

\subsection{Logical Inconsistencies} Logical inconsistencies are those local to the discourse. 
All fallacious mathematical proofs generated by a model are considered as such. 
More generally, all self-contradicting premises in the discourse are logical inconsistencies. 
So, for example, while a model can freely state "\prompt{Alfred Tarski was a Polish-American painter born in Warsaw in 1924}", this would not have logical errors,\footnote{Though it would have factual errors since Alfred Tarski was a logician, born in Warsaw in 1901.} while "\prompt{Kaiser Wilhelm said in 1914 that he would end the war. Sixty years later, in 1914, he would argue that he wanted to end it}" would, since 1914 + 60 is not 1914. 
Common rhetorical devices, such as antitheses ("\prompt{it was the worst of times, it was the best of times}") are not considered logical inconsistencies so long as it is clear they are being used as such.

\subsection{Presence of PII} We consider all names, emails, and websites of existing people related to the subject to be PII. 
Exception to this is when the prompt conditioned the model to output it, for example, by narrating a study carried out by an explicitly-named scientist. 
However, we consider it an inconsistency if it is such that can defame a person. 
An example would be when a person's name is misattributed or inaccurate statements are made about them ("\prompt{You miss 100\% of the shots you don't take, said Albert Einstein}"), and they haven't been mentioned in the prompt. 
Note that the latter example incurs two flaws: factual errors and presence of PII.

\section{Full Breakdown of Results}\label{sec:fullbreakdown}

\subsection{Baseline}\label{sec:baselineapp}

The results for \textsc{baseline}: famous and frequent text openers, is reported in \tabref{baselinetab}. 
Famous and frequent text openers involve both initial sentences of books, and common copy-pasted, or "copy-pasta", text commonly found online ("\prompt{Have you ever heard the tragedy of Darth Plagueis the Wise?}"). %
Our study it is one of the first to incorporate internet culture, such as memes, into the analysis of these models in a context that isn't necessarily harm-based. 

Overall, we found that GPT-3.51 has a much higher-quality output than the other models evaluated, but had memorized content on every single one of the points evaluated. 
In this split, we also fail to find a correlation between memorization and output quality, as evidenced by OPT-IML. 

It is worth noting that BLOOM, ChatGPT, and GPT-3.51 did not output PII. 
BLOOM output tags indicating that PII had been removed, although it had a considerable amount of nonsensical discourse: at some point it interwove text from James Joyce's Ulysses, a real estate ad--without any traceable PII--and an "Eggless Coconut Almond Protein Ice-cream" recipe. 

\begin{center}
\begin{table}[h]
\begin{tabular}{| l || m{0.145\columnwidth} | m{0.1\columnwidth} || m{0.09\columnwidth} | m{0.085\columnwidth} | m{0.085\columnwidth} | m{0.11\columnwidth} |} \hline
Model & Memorized ($\downarrow$) & Original ($\uparrow$) & PII ($\downarrow$) & Logical ($\downarrow$) & Factual ($\downarrow$) & Discourse ($\downarrow$) \\ \hline
BLOOM & \cellcolor{blue!10}\textbf{66.7/40.0}  & \cellcolor{blue!10}\textbf{91.0} & \cellcolor{blue!10}\textbf{0.0} & 60.0 & 20.0 & 60.0 \\ \hline 
ChatGPT & 100.0 / 46.7  & 66.1 & \cellcolor{blue!10}\textbf{0.0} & \cellcolor{blue!10}\textbf{0.0} & \cellcolor{blue!10}\textbf{0.0} & 27.3 \\ \hline 
Galactica & 80.0 / 46.7  & 97.4 & 33.3 & 93.3 & 80.0 & 86.7 \\ \hline 
GPT-3.5 & 93.3 / 80.0  & 62.0 & 6.7 & 13.3 & 13.3 & 46.7 \\ \hline 
GPT-3.51 & 100.0 / 93.3  & 62.4 & \cellcolor{blue!10}\textbf{0.0} & 20.0 & 6.7 & \cellcolor{blue!10}\textbf{13.3} \\ \hline 
GPT-4 & 100.0 / 86.7  & 68.2 & \cellcolor{blue!10}\textbf{0.0} & 20.0 & 13.3 & 33.3 \\ \hline 
LLaMA & 100.0 / 80.0  & 77.3 & \cellcolor{blue!10}\textbf{0.0} & 13.3 & 20.0 & 46.7 \\ \hline 
OPT & 100.0 / 60.0  & 87.3 & 13.3 & 46.7 & 13.3 & 73.3 \\ \hline 
OPT-IML & 100.0 / 60.0  & 90.5 & 20.0 & 60.0 & 26.7 & 80.0 \\ \hline\hline
Total & 95.6 / 65.9  & 78.0 & 8.2 & 36.3 & 21.5 & 51.9 \\ \hline 
\end{tabular}
\caption{Results for the \textsc{baseline} domain, frequent text openers. 
Best performer in bold. 
For memorized text we report the incidence when at least one service flagged the text (left) and as a majority vote (right). 
Overall, we can observe that ChatGPT has the highest-quality discourse by far, although the incidence of memorization is extremely high. 
}\label{tab:baselinetab}
\end{table}
\end{center}

\subsection{Scientific Papers}\label{sec:scienceapp}

The results for our evaluation of the \textsc{scientific papers} category is in \tabref{scitab}. 
This domain involved common titles of scientific articles across various disciplines, following standard formats, such as "\prompt{On the Continuity of}" or "\prompt{Large Language Models Are}". %
Interestingly, for the prompt "\prompt{Importance of Early Intervention For}", all models but GPT-4 and OPT-IML insisted on writing an article on autism, often containing content memorized from existing websites on the subject. 

We found that the incidence of PII injected was greatest, on average, in this category. 
While a grand majority of these issues appeared next to titles, we also found a considerable amount of misattributed work--for example, in-line citations to papers that do exist but do not support the claims being done. 
This latter issue was also metered in the \textsc{factual} category. 
Galactica, GPT-4, OPT-IML and BLOOM both had the highest amount of factual errors, with the first model at some point claiming that the entire Internet was 105 Gb. 
The discourse quality of OPT was also one of the worst. 

Galactica did output the least amount of memorized text and had the second-highest proportion of unique text. %
Nonetheless, the highest-quality discourse was output by GPT-3.51, in spite of memorizing all the points. 

\begin{center}
\begin{table}[h]
\begin{tabular}{| l || m{0.145\columnwidth} | m{0.1\columnwidth} || m{0.09\columnwidth} | m{0.085\columnwidth} | m{0.085\columnwidth} | m{0.11\columnwidth} |} \hline
Model & Memorized ($\downarrow$) & Original ($\uparrow$) & PII ($\downarrow$) & Logical ($\downarrow$) & Factual ($\downarrow$) & Discourse ($\downarrow$) \\ \hline
BLOOM & 86.7 / 33.3  & 86.7 & 26.7 & 33.3 & 40.0 & 66.7 \\ \hline 
ChatGPT & 100.0 / 33.3  & 62.4 & \cellcolor{blue!10}\textbf{0.0} & 23.1 & 38.5 & 30.8 \\ \hline 
Galactica & \cellcolor{blue!10}\textbf{60.0/33.3}  & 92.6 & 20.0 & 26.7 & 40.0 & 73.3 \\ \hline 
GPT-3.5 & 93.3 / 53.3  & 74.4 & 40.0 & 26.7 & 13.3 & 60.0 \\ \hline 
GPT-3.51 & 100.0 / 40.0  & 67.9 & \cellcolor{blue!10}\textbf{0.0} & \cellcolor{blue!10}\textbf{13.3} & 26.7 & \cellcolor{blue!10}\textbf{13.3} \\ \hline 
GPT-4 & 93.3 / 46.7  & 81.4 & 26.7 & 20.0 & 40.0 & 60.0 \\ \hline 
LLaMA & 66.7 / 53.3  & 80.4 & 13.3 & \cellcolor{blue!10}\textbf{13.3} & \cellcolor{blue!10}\textbf{6.7} & 53.3 \\ \hline 
OPT  & 66.7 / 26.7  & \cellcolor{blue!10}\textbf{93.9} & 46.7 & 20.0 & 26.7 & 93.3 \\ \hline 
OPT-IML & 86.7 / 33.3  & 91.6 & 40.0 & 53.3 & 40.0 & 40.0 \\ \hline\hline 
Total  & 85.9 / 39.3  & 81.3 & 25.2 & 25.5 & 30.2 & 54.5 \\ \hline 
\end{tabular}
\caption{Results for the \textsc{scientific papers} domain. 
For memorized text we report the incidence when at least one service flagged the text (left) and as a majority vote (right). 
Overall, it can be seen that the memorization rate was quite high ($86\%$). 
Although Galactica tied with BLOOM, GPT-4, and OPT-IML in terms of factual errors, it returned the lowest amount of memorized text, and had the second-highest proportion of unique text. 
GPT-3.51 had a very high-quality discourse, although all of the points had been flagged as memorized. 
}\label{tab:scitab}
\end{table}
\end{center}

\subsection{Blog Posts}\label{sec:blogapp}

The results for the \textsc{blog post} domain are reported in \tabref{blogtab}. 
This category involves prompts with non-specialized text, such as "\prompt{Visiting York in autumn}", "\prompt{Walnut allergy}", or "\prompt{Lemon meringue pie}". %

Since we did not explicitly prompt ChatGPT to actually write a blog post, but only gave it the prompt and forced it to autocomplete, on multiple instances the model refused to answer. 
We do not count these situations on the final numbers below. 
GPT-3.5 had both the highest amount of memorized output ($100\%$), tied with ChatGPT, and the second-lowest amount of factual errors ($46.7\%$). 
OPT had the second-lowest rate of logical errors ($13.3\%$), and the highest percentage of unique text ($95.5\%$). 
However, it output a considerable amount of names and URLs pointing to existing repositories; and several times output non-English text, along with either pronunciation guides or translations that were inaccurate. 

In terms of factual errors, this category had the largest amount of factual errors. 
We attribute this to the fact that a blog post requires a certain amount of external knowledge when claims are made: for example, a common paragraph structure we observed as "$\langle$Proper Noun$\rangle$ is a $\langle$claim$\rangle$, with $\langle$description$\rangle$. $\langle$Supporting data 1$\rangle$ \dots". 
Given the open-ended nature of the prompts, the models had considerable difficulty outputting factual text, in spite of being one of the categories with the highest memorization rate. 
Nonetheless, the models returned text with an authoritative tone, but counterfactual information. 

For example, for the prompt "\prompt{The Mexican Day of the Dead}", all models correctly described the Day of the Dead as an annual holiday celebrated in Mexico in November 1 and 2. 
However, when expanding on the text, the models started having problems: Galactica described it as a celebration of "the Mexican Revolution and the Mexican national symbol of the day." GPT-3.5 claimed that the holiday was a festival called Mictecacihuatl, but then was brought to Mexico by the Spanish in the 16th century.\footnote{Mictecacihuatl is a goddess, not a festival. The statement is self-contradicting (how can the Spanish bring it if it was already there?), although a claim it made about the catholic All Saint's Day being brought in the 16th century is correct.} 
Other models returned higher quality text with minor (BLOOM) to no (ChatGPT) factual errors for this prompt.

\begin{center}
\begin{table}[h]
\begin{tabular}{| l || m{0.145\columnwidth} | m{0.1\columnwidth} || m{0.09\columnwidth} | m{0.085\columnwidth} | m{0.085\columnwidth} | m{0.11\columnwidth} |} \hline
Model & Memorized ($\downarrow$) & Original ($\uparrow$) & PII ($\downarrow$) & Logical ($\downarrow$) & Factual ($\downarrow$) & Discourse ($\downarrow$) \\ \hline
BLOOM & \cellcolor{blue!10}\textbf{53.3/20.0}  & 92.9 & 20.0 & 53.3 & 73.3 & 86.7 \\ \hline 
ChatGPT & 100.0 / 33.3  & 61.7 & \cellcolor{blue!10}\textbf{0.0} & \cellcolor{blue!10}\textbf{0.0} & 50.0 & 16.7 \\ \hline 
Galactica & 86.7 / 33.3  & 93.6 & 40.0 & 60.0 & 86.7 & 80.0 \\ \hline 
GPT-3.5 & 100.0 / 73.3  & 67.9 & 13.3 & 20.0 & 46.7 & 33.3 \\ \hline 
GPT-3.51 & 93.3 / 73.3  & 66.7 & \cellcolor{blue!10}\textbf{0.0} & \cellcolor{blue!10}\textbf{0.0} & \cellcolor{blue!10}\textbf{26.7} & \cellcolor{blue!10}\textbf{13.3} \\ \hline 
GPT-4 & 100.0 / 66.7  & 75.4 & 13.3 & 13.3 & \cellcolor{blue!10}\textbf{26.7} & 60.0 \\ \hline 
LLaMA & 93.3 / 73.3  & 82.8 & \cellcolor{blue!10}\textbf{0.0} & 33.3 & 46.7 & 73.3 \\ \hline 
OPT & 73.3 / 26.7  & \cellcolor{blue!10}\textbf{95.4} & 33.3 & 13.3 & 53.3 & 86.7 \\ \hline 
OPT-IML & 80.0 / 40.0  & 89.3 & 6.7 & 66.7 & 60.0 & 53.3 \\ \hline\hline 
Total & 88.2 / 48.9  & 80.6 & 14.1 & 28.9 & 52.2 & 55.9 \\ \hline 
\end{tabular}
\caption{Results for the \textsc{blog post} domain. 
For memorized text we report the incidence when at least one service flagged the text (left) and as a majority vote (right). 
This is the category with the highest incidence of factual errors. 
Even then, both GPT-3.51 and ChatGPT output text without PII or logical errors. OPT had the highest amount of original text being output, although the worst (tied with BLOOM) quality of discourse. 
}\label{tab:blogtab}
\end{table}
\end{center}

\subsection{Long Text}\label{sec:longapp}

The results for the \textsc{long text} domain evaluation are reported in \tabref{longtab}. 
This category involves prompts that take text verbatim from well-known corpora, such as speeches and articles. However, unlike the baseline in \secref{baselineapp}, we select the prompts from the middle of the text. 
This allows us to evaluate both incidence of memorization regardless of prompt conditioning (i.e., that the models memorize training data right-to-left), and discourse proficiency under longer prompts. 
Further samples are shown in \secref{samplelongtext}. 

We find that in this category, in aggregate, we obtain the highest-quality discourse, in addition to presenting the second-largest amount of memorized text. 

We also noticed that the models tended to mimic remarkably well the style of the text, both in syntactical and in rhetorical terms. 
For example, an extract from Albert Camus' The Myth of Sisyphus included the citations in IEEE style, and most models would output (fictitious) citations at determined points in the text. 
They were also mostly able to continue capturing the themes in the discourse (e.g., Sisyphus' boulder and the punishment of the gods). 
However, all but two models (ChatGPT and OPT-IML) failed to fully grasp the allegory of the text linking Sisyphus' punishment to the conditions of factory workers, in spite of it being part of the prompt. 
Galactica did remarkably well at the beginning, capturing correctly Camus' analogy; but halfway through the text it mixes up the referents ("the gods are to the proletarian what the enemy is to a hero"); and near the end of the text it re-captures the thesis by mentioning that "his fear concerns the endless cycle of his revolt." 
This text is flagged as having $100\%$ original text, in contrast to the other high-quality output, both GPT-3 models, which were around $30\%$, and GPT-4, which was $2\%$. 

Another example can be seen in the output prompted by William Wilberforce's May 12, 1789 Abolition Speech. All models adopted UK spelling through their outputs, did not space paragraphs (our original prompt is one 543-word passage with a single line break), and returned content in the same style. 
Both BLOOM and GPT-3.51 simply started outputting fake testimonials; in the case of BLOOM to non-existent people, and in the case of GPT-3.51 misattributed to contemporaries of Wilberforce, such as a Mr. William Pitt, "captain of a slave ship" (Pitt the Younger), or a Mr. Rev. Falconbridge (Alexander Falconbridge), who "spent twenty-five years in the Guinea trade." 
Finally, ChatGPT responded by correctly summarizing the prompt's thesis.

\begin{center}
\begin{table}[h]
\begin{tabular}{| l || m{0.145\columnwidth} | m{0.1\columnwidth} || m{0.09\columnwidth} | m{0.085\columnwidth} | m{0.085\columnwidth} | m{0.11\columnwidth} |} \hline
Model & Memorized ($\downarrow$) & Original ($\uparrow$) & PII ($\downarrow$) & Logical ($\downarrow$) & Factual ($\downarrow$) & Discourse ($\downarrow$) \\ \hline
BLOOM & \cellcolor{blue!10}\textbf{26.7/6.7}  & 91.1 & 20.0 & 13.3 & 26.7 & 66.7 \\ \hline 
ChatGPT & 66.7 / 33.3  & 77.4 & \cellcolor{blue!10}\textbf{0.0} & \cellcolor{blue!10}\textbf{0.0} & \cellcolor{blue!10}\textbf{13.3} & \cellcolor{blue!10}\textbf{6.7} \\ \hline 
Galactica & 100.0 / 13.3  & \cellcolor{blue!10}\textbf{97.5} & 33.3 & 60.0 & 80.0 & 86.7 \\ \hline 
GPT-3.5 & 73.3 / 60.0  & 76.4 & 20.0 & 40.0 & \cellcolor{blue!10}\textbf{13.3} & 86.7 \\ \hline 
GPT-3.51 & 73.3 / 60.0  & 74.8 & 26.7 & 33.3 & 53.3 & 53.3 \\ \hline 
GPT-4 & 100.0 / 86.7  & 50.8 & 6.7 & 13.3 & \cellcolor{blue!10}\textbf{13.3} & 20.0 \\ \hline 
LLaMA & 93.3 / 60.0  & 58.9 & \cellcolor{blue!10}\textbf{0.0} & \cellcolor{blue!10}\textbf{0.0} & \cellcolor{blue!10}\textbf{13.3} & 40.0 \\ \hline 
OPT & 46.7 / 20.0  & 91.9 & 26.7 & 40.0 & 46.7 & 60.0 \\ \hline 
OPT-IML & 60.0 / 26.7  & 91.3 & 26.7 & 60.0 & 40.0 & 60.0 \\ \hline\hline 
Total & 71.1 / 40.7  & 78.9 & 17.8 & 28.9 & 33.3 & 53.3 \\ \hline 
\end{tabular}
\caption{Results for \textsc{long text}. For memorized text we report the incidence when at least one service flagged the text (left) and as a majority vote (right). 
This is the (non-baseline) domain with the highest-quality discourse and the second-lowest amount of memorized text. 
Of note is ChatGPT's performance, where in the cases where it did not output memorized text, it tended to output summaries of the prompts. 
BLOOM had the lowest incidence of memorized text. 
Both GPT-3 models had one of the least amounts of memorized text when compared to the other domains, although the discourse quality was not as high as it normally is. 
}\label{tab:longtab}
\end{table}
\end{center}

\subsection{KR}\label{sec:krapp}

The results for the \textsc{KR} domain evaluation are reported in \tabref{krtab}. 
In this domain we evaluate the model's ability to respond correctly to general-knowledge questions ("\prompt{What was the cause of everyone changing residences in New York, on May 1, between 1820 and 1945?}"), as well as explaining its rationale ("\prompt{Is $\exists x\forall y$ the same as $\forall x\exists y$? Give a counterexample.}"). 
We do not perform chain-of-thought prompting. 
Further samples are in \secref{samplekr}. 
In this domain, the amount of factual errors measures the error rate--including when the model does not answer the question. See \secref{samplekr} for this. 

We find that, overall, the models do not do well in this domain. The incidence of memorized text is comparatively lower, and the discourse quality is overall better than in other domains; 
but the factual errors in this category are much higher. 
The incidence of PII across all LLMs is the lowest for all domains, with only four models misattributing work to living scientists and lawyers. 

In terms of discourse, many models (but, in particular, OPT) did not answer the question. 
We also observed that the syntactical structure of the question influenced the response. 
For example, a question designed to spot reasoning errors ("\prompt{My friend Tom says he has a dugite, and the dugite winks at him. Assuming he has a dugite, is Tom lying?}") the models often assumed he was lying for different reasons (e.g., lying about having a dugite, in spite of the assumptions) as opposed to catching the reasoning error (snakes cannot wink). 

In some instances, ChatGPT and GPT-4 would gaslight the user: they would confound their inability to retrieve the answer (such as the question about May 1), with the prompt being a lie. 
They were also be surprisingly good at answering the question, but tended to answer with boilerplate (e.g., "there are a few ways to answer this") followed by an enumeration of possible answers--most of which were wrong. 

OPT-IML, in several instances, was able to correctly answer the question but committed several factual errors when responding. 

One of the models, BLOOM, began to answer the quantum entanglement (\secref{samplekr}) question correctly, but then drifted to make references to the show Quantum Leap, committed a few responsible AI breaches, and then returned to answer the question correctly. 
This phenomenon (lack of cohesive discourse) was consistently present in all but the GPT models.

\begin{center}
\begin{table}[h]
\begin{tabular}{| l || m{0.145\columnwidth} | m{0.1\columnwidth} || m{0.09\columnwidth} | m{0.085\columnwidth} | m{0.085\columnwidth} | m{0.11\columnwidth} |} \hline
Model & Memorized ($\downarrow$) & Original ($\uparrow$) & PII ($\downarrow$) & Logical ($\downarrow$) & Factual ($\downarrow$) & Discourse ($\downarrow$) \\ \hline
BLOOM & \cellcolor{blue!10}\textbf{33.3/13.3}  & 96.1 & 6.7 & 66.7 & 86.7 & 86.7 \\ \hline 
ChatGPT & 73.3 / 40.0  & 86.2 &  \cellcolor{blue!10}\textbf{0.00} & 13.3 & \cellcolor{blue!10}\textbf{80.00} & 40.0 \\ \hline 
Galactica & 60.0 / 6.7  & 78.8 & \cellcolor{blue!10}\textbf{0.00} & 20.0 & \cellcolor{blue!10}\textbf{80.00} & 60.0 \\ \hline 
GPT-3.5 & 73.3 / 13.3  & 84.5 &  \cellcolor{blue!10}\textbf{0.00} & \cellcolor{blue!10}\textbf{0.00} & 86.7 & \cellcolor{blue!10}\textbf{0.00} \\ \hline 
GPT-3.51 & 80.0 / 13.3  & 82.6 & \cellcolor{blue!10}\textbf{0.00} & \cellcolor{blue!10}\textbf{0.00} & 86.7 & \cellcolor{blue!10}\textbf{0.00} \\ \hline 
GPT-4 & 80.0 / 6.7  & 93.2 & \cellcolor{blue!10}\textbf{0.00} & 20.0 & \cellcolor{blue!10}\textbf{80.00} & 6.7 \\ \hline 
LLaMA & 60.0 / 26.7  & 91.3 & 20.0 & 26.7 & 93.3 & 66.7 \\ \hline 
OPT & 40.0 / 0.0  & 96.5 & 6.7 & 80.0 & 100.0 & 80.0 \\ \hline 
OPT-IML & 46.7 / 13.3  & \cellcolor{blue!10}\textbf{96.7} & 13.3 & 33.3 & 100.0 & 80.0 \\ \hline\hline 
Total & 60.0 / 14.8  & 91.3 & 5.2 & 28.9 & 88.2 & 46.7 \\ \hline
\end{tabular}
\caption{Results for the \textsc{KR} domain on the discourse analysis part of our work. 
For memorized text we report the incidence when at least one service flagged the text (left) and as a majority vote (right). 
The error rate is captured in the factual column, and it also considers not answering the question a factual error. 
Overall the models had very low accuracy, although the proportion of unique text and discourse quality was generally better than in other categories. 
}\label{tab:krtab}
\end{table}
\end{center}

\section{Sample Discourse Analysis}\label{sec:samplesyntax}

In this section we show a sample of our discourse evaluation process. Consider the following prompt from \textsc{scientific papers}: \prompt{The Alaskan Malamute}. 
The output obtained by Galactica is (prompt underlined; skipped and removed content in braces):

\vspace{10pt}
\begin{quote}
\small{
    \underline{The Alaskan Malamute} \cellcolor{blue!10}\textbf{(Alopex lagopus)} and Siberian Husky \cellcolor{blue!10}\textbf{(E. h. laniger)} are \cellcolor{blue!10}\textbf{medium-sized (108 to 210 kg, 240 to 460 lb) husky dogs} bred for sledding over rough terrain in cold weather. The Alaskan Malamute originated on St. Lawrence Island and the Alaskan coast during the \{(removed term)\} era. It is a powerful, muscular breed. The Alaskan has one of the largest, most rounded heads and broad chests of any husky. A hardy, tough-natured breed, it is one of the oldest sled dog breeds. It originated in the 1800s and was used by the \{(removed term)\} to hunt and herd caribou. These dogs were originally bred for survival at high altitudes and harsh conditions.
    
\{(skipped content)\}

The Labrador retriever is a medium-sized (40 to 70 kg, 88 to 154 lb) purebred dog developed in the United Kingdom in the 19th century. These well-developed dogs have a compact, balanced appearance. They are a popular companion dog, being the favorite of many who look for a loyal companion as much as someone who enjoys the physical activity that the Labrador makes a good athlete for. Labradors make excellent family pets, but they should not be housed with smaller animals, as their size could be a danger to them. Labradors are active dogs that thrive on a wide variety of outside activities.

The Golden retriever is a \cellcolor{blue!10}\textbf{medium-sized (35 to 70 kg, 77 to 154 lb)} dog breed that originates in the United States. They are an extremely popular dog for active individuals who want an intelligent and loving family pet that can be kept indoors for long hours without much trouble. Golden retrievers are extremely intelligent, obedient, eager to please, and easy to train. They are known for their \cellcolor{blue!10}\textbf{great stamina and athleticism} and are considered a highly responsive breed.\cellcolor{blue!10}\textbf{ An active playful dog}, \cellcolor{blue!10}\textbf{the Golden retriever is the perfect companion for those who} have an active lifestyle but \cellcolor{blue!10}\textbf{just aren’t into high-energy dogs} that like to be outside.
}
\end{quote}
\vspace{10pt}

This text presents several flaws, bolded in the text. For example:

\begin{itemize}
    \item \textsc{Logical}: In the first paragraph, medium-sized is between $108-210$ kg, but in the next two, medium-sized is $35-70$ and $40-70$ kg. It states that Golden Retrievers are active playful dogs and then proceeds to point out that they can stay indoors, when a $70$ kg active and playful dog probably would require a lot of space. 
    \item \textsc{Factual}: The taxonomy of both the Alaskan Malamute and the Siberian Husky is wrong--\emph{alopex lagopus} is the arctic fox, and \emph{laniger} is the species name for the eastern woolly Lemur. Weights of the Alaskan Malamute are wrong, as no dog is able to reach $460$ lb. The model considers both to be husky dogs, in spite of being different breeds. In the first paragraph alone, there are further counterfactual statements around the purpose, location, and time of their breeding.
    \item \textsc{Discourse}: Presence of terms considered pejorative by certain populations (not shown in the text).
    \item \textsc{PII}: No PII errors are found. 
\end{itemize}

Three of the services flagged existing memorized content, to an average of $93\%$ unique text. 

\section{Sample Prompts}\label{sec:sampleprompts}

\subsection{KR}\label{sec:samplekr}

Here is a sample prompt from the \textsc{KR} task. 
\begin{quote}
\begin{sloppypar}
\prompt{Suppose I give you one-half of an entangled qubit \textbackslash ket\{00\} + \textbackslash ket\{11\} and I take the other half. Then I take a faster-than-light spaceship and go to a place 3 light years away--that is, the speed of light takes 3 years to get to you. 
Now suppose I measure my qubit, and obtain \textbackslash ket\{0\}. 
How fast would it take you to notice that your state has collapsed, and to what?}
\end{sloppypar}
\end{quote}
Remarkably, several models--such as GPT-3.5 and GPT-4--were able to respond this correctly.

\subsection{Long Text Prompt}\label{sec:samplelongtext}

Here is a sample prompt from the \textsc{long text} category. 
It belongs to the "We shall fight on the beaches" speech given in 4 June 1940, by Winston Churchill. 
\begin{quote}
\begin{sloppypar}
\prompt{The French High Command hoped they would be able to close the gap, and the Armies of the north were under their orders. Moreover, a retirement of this kind would have involved almost certainly the destruction of the fine Belgian Army of over 20 divisions and the abandonment of the whole of Belgium. Therefore, when the force and scope of the German penetration were realized and when a new French Generalissimo, General Weygand, assumed command in place of General Gamelin, an effort was made by the French and British Armies in Belgium to keep on holding the right hand of the Belgians and to give their own right hand to a newly created French Army which was to have advanced across the Somme in great strength to grasp it.}
\end{sloppypar}
\end{quote}
\begin{quote}
\begin{sloppypar}
\prompt{However, the German eruption swept like a sharp scythe around the right and rear of the Armies of the north. Eight or nine armored divisions, each of about four hundred armored vehicles of different kinds, but carefully assorted to be complementary and divisible into small self-contained units, cut off all communications between us and the main French Armies. It severed our own communications for food and ammunition, which ran first to Amiens and afterwards through Abbeville, and it shore its way up the coast to Boulogne and Calais, and almost to Dunkirk. Behind this armored and mechanized onslaught came a number of German divisions in lorries, and behind them again there plodded comparatively slowly the dull brute mass of the ordinary German Army and German people, always so ready to be led to the trampling down in other lands of liberties and comforts which they have never known in their own.}
\end{sloppypar}
\end{quote}
\begin{quote}
\begin{sloppypar}
\prompt{I have said this armored scythe-stroke almost reached Dunkirk-almost but not quite. Boulogne and Calais were the scenes of desperate fighting. The Guards defended Boulogne for a while and were then withdrawn by orders from this country. The Rifle Brigade, the 60th Rifles, and the Queen Victoria’s Rifles, with a battalion of British tanks and 1,000 Frenchmen, in all about four thousand strong, defended Calais to the last. The British Brigadier was given an hour to surrender. He spurned the offer, and four days of intense street fighting passed before silence reigned over Calais, which marked the end of}
\end{sloppypar}
\end{quote}

\bibliographystyle{elsarticle-harv}
\bibliography{biblio}

\end{document}